\documentclass[10pt,twocolumn,letterpaper]{article}

\usepackage[pagenumbers]{cvpr}

\makeatletter
\@namedef{ver@everyshi.sty}{}
\makeatother

\usepackage{graphicx}
\usepackage{amsmath}
\usepackage{amssymb}
\usepackage{booktabs}

\usepackage{multicol}
\usepackage{multirow}
\usepackage{algorithm}
\usepackage{algorithmic}

\usepackage{xcolor}
\usepackage[pagebackref,breaklinks,colorlinks,bookmarks=false,urlcolor=black,citecolor={green!60!black}]{hyperref}

\usepackage{colortbl}
\definecolor{cGreen}{RGB}{0,255,0}
\definecolor{LightCyan}{rgb}{0.88,1,0.88}
\definecolor{bananamania}{rgb}{0.98, 0.91, 0.71}

\usepackage{makecell}

\usepackage{multibib}
\newcites{latex}{References}

\usepackage{tcolorbox}
\usepackage{comment}

\usepackage{dsfont}
\newcommand{\mIdent}{\mathbf{{I}}}

\DeclareMathOperator*{\argmin}{arg\,min}

\usepackage[capitalize]{cleveref}
\crefname{section}{Sec.}{Secs.}
\Crefname{section}{Section}{Sections}
\Crefname{table}{Table}{Tables}
\crefname{table}{Tab.}{Tabs.}

\newcommand{\mycomment}[1]{}

\makeatletter
\def\ps@myheadings{%
    \let\@oddfoot\@empty\let\@evenfoot\@empty
    \def\@evenhead{\thepage\hfil\slshape\leftmark}%
    \def\@oddhead{{\slshape\rightmark}\hfil\thepage}%
    \let\@mkboth\@gobbletwo
    \let\sectionmark\@gobble
    \let\subsectionmark\@gobble
    }
  \if@titlepage
  \renewcommand\maketitle{\begin{titlepage}%
  \let\footnotesize\small
  \let\footnoterule\relax
  \let \footnote \thanks
  \null\vfil
  \vskip 60\p@
  \begin{center}%
    {\LARGE \@title \par}%
    \vskip 3em%
    {\large
     \lineskip .75em%
      \begin{tabular}[t]{c}%
        \@author
      \end{tabular}\par}%
      \vskip 1.5em%
    {\large \@date \par}
  \end{center}\par
  \@thanks
  \vfil\null
  \end{titlepage}%
  \setcounter{footnote}{0}%
}
\else
\renewcommand\maketitle{\par
  \begingroup
    \renewcommand\thefootnote{\@fnsymbol\c@footnote}%
    \def\@makefnmark{\rlap{\@textsuperscript{\normalfont\color{black}\@thefnmark}}}%
    \long\def\@makefntext##1{\parindent 1em\noindent
            \hb@xt@1.8em{%
                \hss\@textsuperscript{\normalfont\@thefnmark}}##1}%
    \if@twocolumn
      \ifnum \col@number=\@ne
        \@maketitle
      \else
        \twocolumn[\@maketitle]%
      \fi
    \else
      \newpage
      \global\@topnum\z@   
      \@maketitle
    \fi
    \thispagestyle{plain}\@thanks
  \endgroup
  \setcounter{footnote}{0}%
}
\makeatother

\def\cvprPaperID{9145} 
\def\confName{CVPR}
\def\confYear{2023}

\usepackage{txfonts}
\newcommand{\heart}{\ensuremath\varheartsuit}

\begin{document}

\title{Learning Partial Correlation based Deep Visual Representation for Image Classification}

\author{Saimunur Rahman$^{1,2}$, Piotr Koniusz\thanks{Corresponding author. $\quad$This paper is published at CVPR 2023. $\quad$The code is available at   \href{https://github.com/csiro-robotics/iSICE}{https://github.com/csiro-robotics/iSICE} }$^{\;\;,1,3}$, Lei Wang$^2$, Luping Zhou$^4$, Peyman Moghadam$^{1,5}$, Changming Sun$^1$\\
$^1$Data61\heart CSIRO, $^2$University of Wollongong, $^3$Australian National University, \\ $^4$University of Sydney, $^5$Queensland University of Technology 
\\
{\tt\small name.surname@data61.csiro.au, leiw@uow.edu.au, luping.zhou@sydney.edu.au}
}
\maketitle

\begin{abstract}
Visual representation based on covariance matrix has demonstrates its efficacy for image classification by characterising the pairwise correlation of different channels in convolutional feature maps. However, pairwise correlation will become misleading once there is another channel correlating with both channels of interest, resulting in the ``confounding'' effect. For this case, ``partial correlation'' which removes the confounding effect shall be estimated instead. Nevertheless, reliably estimating partial correlation requires to solve a symmetric positive definite matrix optimisation, known as sparse inverse covariance estimation (SICE). How to incorporate this process into CNN remains an open issue. 
In this work, we formulate SICE as a novel structured layer of CNN. To ensure  end-to-end trainability, we develop an iterative method to solve the above matrix optimisation during forward and backward propagation steps. Our work obtains a partial correlation based deep visual representation and mitigates the small sample problem  often encountered by covariance matrix estimation in CNN. 
Computationally, our model can be effectively trained with GPU and works well with a large number of channels of advanced CNNs. Experiments show the efficacy and  superior classification performance of our deep visual representation compared to covariance matrix based counterparts.
\end{abstract}

\vspace{-0.5cm}
\section{Introduction}
Learning effective visual representation is a central issue in computer vision. In the past two decades, describing images with local features and pooling them to a global representation has shown promising performance. As one of the pooling methods, covariance matrix based pooling has attracted much attention due to its exploitation of second-order correlation information of features. A variety of tasks such as fine-grained image classification \cite{lin2015bilinear}, image segmentation \cite{ionescu2015matrix}, generic image classification \cite{li2017second,lin2018second,rahman2020redro}, image set classification \cite{wang2012covariance}, action recognition \cite{koniusz2021tensor}, few-shot classification \cite{zhang2019few} and few-shot detection \cite{zhang2020sopaccv,zhang2022kernelized,zhang2022time} have benefited from the covariance matrix based representation. A few pioneering works have integrated covariance matrix as a pooling method within convolutional neural networks (CNN) and investigated associated issues such as matrix function backpropagation \cite{ionescu2015matrix}, matrix normalisation \cite{lin2017bilinear,li2018towards,song2021approximate}, compact matrix estimation \cite{gao2016compact,yu2022fast} and kernel based extension \cite{engin2018deepkspd}. The above works further improved visual representations based on covariance matrix.

\begin{figure}[t]
\vspace{-0.3cm}
  \centering
  \includegraphics[clip, trim=0cm 0cm 0cm 0cm, width=0.49\textwidth]{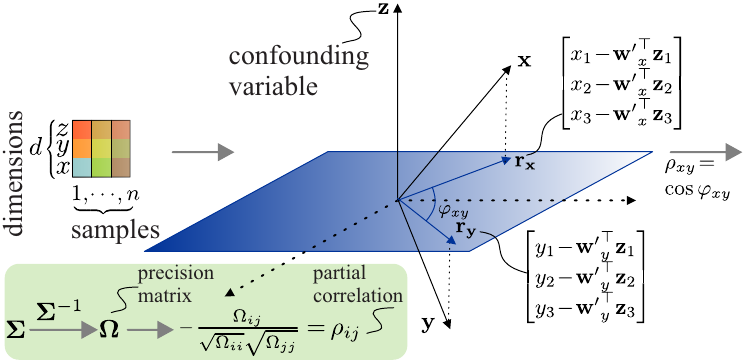}
  \caption{Understanding the partial correlation (a  3D toy case). Unlike the ordinary covariance (pairwise correlation of say $\mathbf{x}$ and $\mathbf{y}$ corresponding to channels), partial correlation between variables $\mathbf{x}$ and $\mathbf{y}$  removes the influence of the confounding variable $\mathbf{z}$. Let the number of samples  $n\!=\!3$ and channels $d\!=\!3$. For the 3D case,  $\mathbf{x}$ and $\mathbf{y}$ are projected onto a {plane} perpendicular to $\mathbf{z}$. Then $\rho_{xy}\!=\!\cos{\varphi_{xy}}$ (and $\rho_{xz}$ and $\rho_{yz}$ can be computed by analogy). Projected ``residuals'' $\mathbf{r}_\mathbf{x}$ and $\mathbf{r}_\mathbf{y}$ are computed as indicated in the plot,   ${\mathbf{w}'\!}_x\!=\!\argmin_{\mathbf{w}}\sum_{i=1}^3(x_i\!-\!\mathbf{w}_x^\top\mathbf{z}_i)$ where $\mathbf{z}_i\!=\![z_i, 1]^\top$ (and ${\mathbf{w}'\!}_y$ is computed by analogy). The green box: for $d\!>\!3$, the computation of partial correlation requires covariance inversion \cite{GVK021834997}.}
  \label{fig:conf}
\end{figure}

\begin{figure*}
    \centering
    \includegraphics[width=\textwidth]{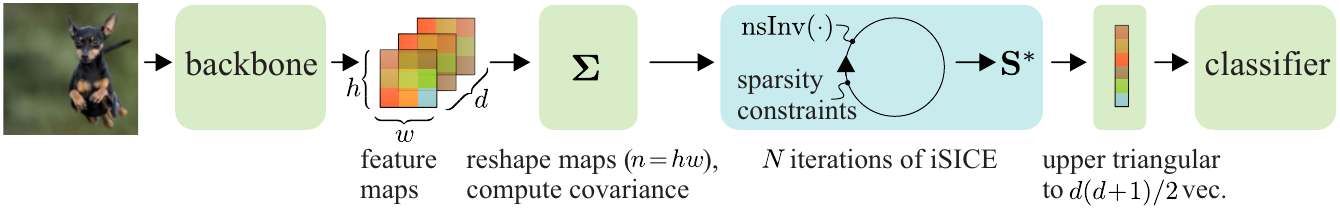}
    \caption{Proposed iterative sparse inverse covariance estimation (iSICE) method in a CNN pipeline.}
    \label{fig: method}
\end{figure*}

Despite the above progress, covariance matrix merely measures the pairwise correlation (more accurately, covariance) of two variables \textit{without} taking any other variables into account. This can be easily verified because its $(i,j)$-th entry solely depends on the $i$-th and $j$-th variables on a sample set. In Statistics, it is known that such a pairwise correlation will give misleading results once a third variable is correlated with both variables of interest due to the ``confounding'' effect. For this situation, the partial correlation is the right measure to use. It regresses out the effects of other variables from the two variables and then calculates the correlation of their residuals instead. Partial correlation can be conveniently obtained by computing inverse covariance matrix, also known as the precision matrix \cite{GVK021834997} in the statistical community. Figure \ref{fig:conf} illustrates the geometrical interpretation of partial correlation.

The above observation motivates us to investigate a visual representation for image classification based on the inverse covariance matrix. After all, it has better theoretical support on {characterising} the essential relationship of variables (\eg, the channels in a convolutional feature map) when other variables are present. {Note that inverse covariance matrix can be used for many vision tasks but in this paper, we investigate it from the perspective of image classification.} Nevertheless, reliably estimating inverse covariance matrix from the local descriptors of a CNN feature map is a challenging task. This is primarily due to the small spatial size of the feature map, \ie, sample size, and a higher number of channels, \ie, feature dimensions, and this issue becomes more pronounced for advanced CNN models. An unreliable estimate of inverse covariance matrix will critically affect its effectiveness as a visual representation. 
One might argue that by increasing the size of input images or using a dimension reduction layer to reduce the number of feature channels, such an issue could be resolved. In this paper, we investigate this issue  from the perspective of robust precision matrix estimation.

To achieve our goal, we explore the use of sparsity prior for inverse covariance matrix estimation in the literature. Specifically, the general principle of ``bet on sparsity'' \cite{hastie2009elements} is adopted in estimating the structure of high-dimensional data, and this leads to an established technique called sparse inverse covariance estimation (SICE) \cite{friedman2008sparse}. It solves an optimisation in the space of symmetric and positive definite (SPD) matrix to estimate the inverse covariance matrix by imposing the sparsity prior on its entries. SICE is designed to handle small sample problem and it is known for its excellent effectiveness to that end \cite{friedman2008sparse}. An initial attempt to apply SICE for visual representation is based on handcrafted or pre-extracted features of small size and an off-the-shelf SICE solver, and it does not have the ability to backpropagate through SICE due to optimisation of the SPD matrix with the imposed non-smooth  sparsity term \cite{zhang2021beyond}.

Our work is the first one that truly integrates SICE into CNN for end-to-end training. Clearly, such an integration will fully take advantage of the feature learning capability of CNN and the partial correlation offered by inverse covariance matrix. On the other hand, realising such an integration is not trivial. Unlike covariance matrix, which is obtained by simple arithmetic operations, SICE is obtained by solving an SPD matrix based optimisation. How to incorporate this optimisation process into CNN as a layer is an issue. Furthermore, this SICE optimisation needs to be solved for each training image during both forward and backward phases to generate a visual representation. Directly solving this optimisation within CNN will not be practical even for a medium-sized SICE problem.

To efficiently integrate SICE into CNN, we propose a fast end-to-end training method for SICE by taking inspiration from Newton-Schulz iteration \cite{higham2008functions}. Our method solves the SICE optimisation with a smooth convex cost function by re-parameterising the non-smooth term in the original SICE cost function (see Eq. (\ref{eq: sice_obj})), and it can therefore be optimised with standard optimisation techniques such as gradient descend. Furthermore, we effectively enforce the SPD constraint during optimisation so that the obtained SICE solution remains SPD as desired. Figure \ref{fig: method} shows our ``Iterative Sparse Inverse Covariance Estimation (iSICE)''. In contrast to SICE, iSICE works with end-to-end trainable deep learning models.
Our iSICE  involves simple matrix arithmetic operations fully compatible with GPU. It can approximately solve large SICE problems within CNN efficiently.

\vspace{0.2cm}
Our main contributions are summarised as follows.
\begin{enumerate}
    \item To more precisely characterise the relationship of features for visual representation, this paper proposes to integrate sparse inverse covariance estimation (SICE) process into CNNs as a novel layer. To achieve this, we develop a method based on Newton-Schulz iteration and box constraints for $\ell_1$ penalty to solve the SICE optimisation with CNN and maintain the end-to-end training efficiency. To the best of our knowledge, our iSICE is the first end-to-end SICE solution for CNN. 
    \item Our iSICE method requires a minimal change of network architecture. Therefore, it can readily be integrated with existing works to replace those using deep network models to learn covariance matrix based visual representation. The iSICE is fully compatible with GPU and can be easily implemented with modern deep learning libraries.
    \item As the objective of SICE is a combination of $\log\det$ term (may change rapidly) and sparsity (changes slowly), achieving the balance between both terms during optimisation by the gradient descent is hard. To this end, we propose a minor contribution: a simple modulating network whose goal is to adapt on-the-fly learning rate and sparsity penalty.
\end{enumerate}

Experiments on multiple image classification  datasets show the effectiveness of our proposed iSICE method. 

\section{Related work}
Since the advent of covariance representation methods in deep learning  \cite{lin2015bilinear,gao2016compact}, reliable estimation of the covariance matrix from a CNN feature map remains an issue. The issue is due to the small spatial size of feature map (corresponding to the number of samples) and the higher number of feature channels (corresponding to the feature dimensions), which could cause unreliable estimation and even matrix singularity due to the so-called ``curse of dimensionality''. Existing works either append a small positive constant to the diagonals of covariance matrix \cite{lin2015bilinear} or use matrix normalisation operation \cite{ionescu2015matrix,lin2017bilinear,li2017second} to handle this issue. Matrix normalisation approaches mitigate unbalanced  spectrum represented by eigenvalues of covariance matrix \cite{s17,li2017second,lin2017improved,kon_tpami2020a,sfa_yifei}. Different from the existing methods, in this paper, we approach the reliable covariance estimation problem in CNN from a perspective of partial correlations that can be efficiently captured by the inverse of covariance matrix, also known as the precision matrix \cite{GVK021834997}, when a large number of samples is available for estimation.

Covariance representation strives to capture the underlying structure of CNN feature channels. 
In the literature of knowledge representation \cite{kubat1999neural}, it is recommended to leverage prior knowledge to improve a learning task when sufficient data is not available\footnote{This is indeed the case with CNNs as the spatial size of a CNN feature map is usually small when compared to the number of its channels.}. Thus, prior knowledge can be used to improve the estimation of underlying structure of high-dimensional data captured by covariance representation. One common prior knowledge is ``structure sparsity'' which leads to the sparse inverse covariance matrix in the literature of statistical machine learning \cite{friedman2008sparse,huang2011learning}. 

Structure sparsity cannot be readily applied to covariance representation as it requires the access to partial correlation between feature components. A covariance matrix captures pairwise correlation of feature components without 
taking into account the confounding effect of remaining components. 
Therefore, it is unlikely that the covariance matrix will be sparse by nature. To obtain partial correlation, SICE moves from covariance matrix to its inverse. An inverse covariance matrix captures partial correlation between feature components by regressing out the effects of other variables \cite{huang2011learning}. Once other variables are factored out, structure sparsity can be effectively enforced in SICE. 
In a recent work \cite{zhang2021beyond}, SICE-based visual representation has been applied to image classification with handcrafted and pre-extracted  features of small size. However, SICE has never been integrated into CNN for end-to-end training with the goal of adapting to such a representation. The existing solvers for computing SICE also have limited GPU support \cite{friedman2008sparse,diamond2016cvxpy}. Thus, we propose an end-to-end trainable iterative method for solving SICE optimisation with CNNs.

\section{Proposed method}
In this section, we begin by discussing the background of SICE. Then we discuss how it can be estimated from CNN feature descriptors. Finally, we describe our proposed method which is trainable end-to-end with a CNN.

\subsection{The basic idea of SICE}
As a representation, the covariance matrix captures the underlying structure of a feature set. It uses a covariance matrix estimated from samples to capture this structure. Sparse inverse covariance
estimation (SICE) focuses on the following two issues: (1) instability or singularity of sample-based covariance matrix estimated from a small number of high-dimensional feature vectors. This situation makes it less effective in capturing the underlying structure of data. As an example, in this case the smaller and larger eigenvalues of the estimated covariance matrix become poorly estimated. Thus, a suitable regularisation (a prior) is needed  during the estimation to mitigate the bias of our estimator; (2) rigid estimation of covariance matrix for high-dimensional feature vectors is not always appropriate as the high-dimensional data usually presents a complex structure. If there is prior knowledge available, it should be used to improve the covariance estimation from a small number of samples.

In terms of prior knowledge of high-dimensional data, structure sparsity \cite{huang2011learning} and the ``bet on sparsity'' principle \cite{hastie2009elements} are the two common  priors used in the literature. Suppose that we have a probabilistic graphical model, where each node corresponds to a feature and the statistical dependence between two nodes is expressed with an edge linking two nodes. Structure sparsity would specify how sparse such a graph is, \eg, how many edges are presented in this graph. More generally, even if there is no clear prior knowledge on structure sparsity available, the ``bet on sparsity'' principle can still be applied to estimate the structure of the graph by imposing a sparsity prior. Its rationality is as follows. If the graph is indeed sparse, SICE will estimate its underlying structure with a correct prior, and if the graph is dense, it will not estimate the underlying structure accurately with such a prior. However, in the latter case, we will not lose much because we have known that we do not have enough sample to estimate the dense structure. The ``bet on sparsity'' principle has been widely adopted in high-dimensional data analysis, and also demonstrated its efficacy on covariance matrix based visual representation estimated from handcrafted or pre-extracted CNN features \cite{zhang2021beyond}.

SICE strives to improve the covariance estimation with the use of prior knowledge. To incorporate prior knowledge, SICE switches to the inverse of covariance matrix from covariance matrix. In principle, the covariance matrix captures the apparent pairwise correlation between feature components, \ie, indirect correlation. In comparison, the inverse of covariance matrix is able to characterise the direct (\ie partial) correlation between two feature components by regressing out the remaining features. Using the inverse covariance matrix not only helps to interpret the essential relationship between two features, but also allows the convenient incorporation of the sparsity prior.  

\subsection{Our SICE estimation with CNN}
Suppose, we have the sample-based covariance matrix $\mathbf{\Sigma}$ computed from a set of CNN local descriptors presented in a convolutional feature map. Let $\mathbf{S}$ denote the corresponding sparse inverse covariance matrix.  The off-diagonal entries of $\mathbf{S}$ capture the direct correlation between different descriptor components. They are zero if two components are independent 
under the removed influence of confounding variables. In the literature \cite{friedman2008sparse}, the estimation of $\mathbf{S}$ has been effectively resolved by the maximization of a penalised log-likelihood of data with an SPD constraint on $\mathbf{S}$ and the sparsity prior to induce sparse graph connectivity. The optimal solution of the above problem is known as SICE.
\definecolor{beaublue}{rgb}{0.8, 0.85, 0.8}
\definecolor{blackish}{rgb}{0.2, 0.2, 0.2}
\begin{tcolorbox}[grow to left by=0.01\linewidth, width=1\linewidth, colframe=blackish, colback=beaublue, boxsep=0mm, arc=2mm, left=2mm, right=2mm, top=2mm, bottom=2mm]
\textbf{SICE} is defined as follows:
\begin{equation}
    \mathbf{S}^* = \mathrm{arg}\, \underset{\mathbf{S}\succ 0}{\mathrm{max}}\;\; \mathrm{log}\,\mathrm{det}(\mathbf{S}) - \mathrm{trace}(\mathbf{\Sigma S})-\lambda\lVert\mathbf{S}\rVert_1,
\label{eq: sice_obj}
\end{equation}
where $\mathbf{\Sigma}$ is a sample-based covariance matrix, and $\mathrm{det}(\cdot)$, $\mathrm{trace}(\cdot)$ and $\lVert\cdot\rVert_1$ denote the determinant, trace and $\ell_1$-norm of a vectorization of matrix, respectively.
\end{tcolorbox}
To obtain reliable and faithful SICE, the term $\lVert\mathbf{S}\rVert_1$ imposes the structure sparsity on $\mathbf{S}$. $\lambda$ controls the trade-off between the amount of sparsity and the log-likelihood estimation. The  problem in Eq. \eqref{eq: sice_obj} is convex and can be solved by the off-the-shelf packages such as GLASSO \cite{friedman2008sparse} and CVXPY \cite{diamond2016cvxpy}. However, the objective is non-smooth due to the $\ell_1$ penalty. The above optimisation packages cannot be  used with CNN layers to conduct training with backpropagation. A recent extension of CVXPY called CVXPYLAYERS \cite{agrawal2019differentiable} provides differentiable optimisation layers. However, based on our investigation, it has the following issues: (1) it cannot efficiently solve large SICE problems, \ie, of size $128\!\times\!128$ or higher; (2) it relies on multiple CPU based libraries including CVXPY to solve the optimisation problem and obtain gradients for backpropagation. This greatly limits its efficiency due to the lack of GPU support. The above limitations motivate us to develop an SICE method suitable for end-to-end training with GPU.

\subsection{Proposed end-to-end trainable SICE method}
Let $J$ be the objective function of Eq. \eqref{eq: sice_obj}. $J$ can be optimised by taking the gradient with respect to $\mathbf{S}$ as follows:
\begin{align}
    \frac{\partial J}{\partial \mathbf{S}} &= \frac{\partial}{\partial \mathbf{S}}\log\,\det(\mathbf{S}) - \frac{\partial}{\partial \mathbf{S}} \mathrm{trace}(\mathbf{\Sigma S})- \lambda \frac{\partial}{\partial \mathbf{S}} \lVert\mathbf{S}\rVert_1 \nonumber \\
    &=\mathbf{S}^{-1} - \mathbf{\Sigma} - \lambda\Big(\frac{\partial }{\partial \mathbf{S}} \mathbf{S}^+ -\frac{\partial}{\partial \mathbf{S}} \mathbf{S}^-\Big)\nonumber \\
    &= \mathbf{S}^{-1} - \mathbf{\Sigma} - \lambda\big(\mathrm{sign}(\mathbf{S}^+)-\mathrm{sign}(\mathbf{S}^-)\big), 
    \label{eq: isice}
\end{align}
where $\mathbf{S}^+\equiv\mathrm{max}(0,\mathbf{S})$ and $\mathbf{S}^-\equiv\mathrm{max}(0,-\mathbf{S})$ contain the positive and negative parts of $\mathbf{S}$, respectively. Eq. \eqref{eq: isice} can be optimised with the projected gradient descend which has the native backpropagation support on GPU and can take advantage of GPU parallel computing to improve their speed. Now we discuss how Eq. \eqref{eq: isice} can be effectively optimised using a few consecutive structured CNN layers.

The overview of our method is given in Fig. \ref{fig: method}. From the left, we pass an input image to the backbone and process it till the last convolution layer. We obtain the feature map, \ie, $h\times w\times d$ tensor, where $h$ is the height, $w$ is the width, and $d$ is the number of channels. By reshaping the feature map to a set of $n$  vectors of length $d$, where $n=wh$ and stacking them as column vectors, we can create a $d\times n$ data matrix $\mathbf{X}$. A sample-based covariance matrix $\mathbf{\Sigma}$ estimated from $\mathbf{X}$ is defined as $\mathbf{\Sigma}\equiv\mathbf{X\bar{I}}\mathbf{X}^{\top}$, where $\mathbf{\bar{I}} = \frac{1}{n}(\mathbf{I}-\frac{1}{n}\mathbf{1}\mathbf{1}^\top)$ performs centering of matrix $\mathbf{X}$, where $\mathbf{I}$ and $\mathbf{1}\mathbf{1}^\top$ are $n\times n$ dimensional identity matrix and matrix of all-ones, respectively. Below, we describe  key steps of our method. 

\vspace{0.1cm}
\noindent\textbf{Estimation of precision matrix $\mathbf{S}_0\!=\!\mathbf{\Sigma}'^{-1}$.}
Newton-Schulz iteration is popular as it can approximate the matrix square root\footnote{Notice we use Newton-Schulz iteration to obtain the precision matrix $\mathbf{\Omega}$ from covariance $\mathbf{\Sigma}$. We do not propose or use square rooting of $\mathbf{\Sigma}$, but we compare our iSICE to this kind of covariance normalisation.} fast on GPU \cite{li2018towards}. 
In contrast, during the estimation of $\mathbf{S}$, we use  Newton-Schulz iteration \cite{higham2008functions} for a fast approximate inverse of matrix, 
which imposes a convergence condition $\lVert\mathbf{\Sigma}-\mathbf{I}\rVert_2\!<\!1$ on Algorithm \ref{alg: inv sqrt}. Thus, we normalise $\mathbf{\Sigma}$ by its trace and use the trace-normalised  $\mathbf{\Sigma}'\!=\mathbf{\Sigma}/\mathrm{trace}(\mathbf{\Sigma})$.
Then, the square of the inverse square root $\mathbf{Q}$ is post-normalized by the trace to reverse it, \ie, $\mathbf{\Sigma}^{-1}\!=\!\mathbf{Q}\mathbf{Q}^\top/\mathrm{trace}(\mathbf{\Sigma})$.

 As Eq. \eqref{eq: isice} has to start with an initial $\mathbf{S}_0$, if $\mathbf{\Sigma}'$ is invertible,  Alg. \ref{alg: inv sqrt}  approximates its inverse\footnote{Note that the conventional matrix inverse requires matrix eigendecomposition which is not well supported on GPU \cite{ionescu2015matrix}. In contrast, Newton-Schulz iteration \cite{higham2008functions} is known for fast convergence to  $\mathbf{\Sigma}^{-1}$.}. Although   it is an approximation, we denote it in Alg. \ref{alg: inv sqrt} as $\mathbf{\Sigma}^{-1}$ for brevity. 
 
\mycomment{
\begin{algorithm}[t]
\caption{Fast approximate matrix inverse via Newton-Schulz iterations (named $\text{nsInv}(\cdot)$ in short).}
\label{alg: inv sqrt}
\textbf{Input}: Covariance matrix $\mathbf{\Sigma}$, number of iterations N.\\
\textbf{Output}: Inverse covariance matrix $\mathbf{\Sigma}^{-1}$.

\begin{algorithmic}[1]
\STATE $\mathbf{\Sigma}'$ = $\mathbf{\Sigma}$/trace($\mathbf{\Sigma}$)
\STATE $\mathbf{P} = \frac{1}{2}(3\mathbf{I}-\mathbf{\Sigma}')$, $\mathbf{Y}_0 = \mathbf{\Sigma}'\mathbf{P}$, $\mathbf{Z}_0 = \mathbf{P}$
\FOR{$i=1$ to $ N$} 
    \STATE $\mathbf{P} = \frac{1}{2}(3\mathbf{I}-\mathbf{Z}_{i-1}\mathbf{Y}_{i-1})$\;
    \STATE $\mathbf{Y}_i = \mathbf{Y}_{i-1}\mathbf{P}$, $\mathbf{Z}_i = \mathbf{P}\mathbf{Z}_{i-1}$\;
\ENDFOR
\STATE $\mathbf{Q} = \frac{1}{2}(3\mathbf{I}-\mathbf{Z}_{{N}}\mathbf{Y}_{{N}})\mathbf{Z}_{{N}}$
\STATE $\mathbf{\Sigma}^{-1} = \mathbf{Q}\mathbf{Q}^\top/\mathrm{trace}(\mathbf{\Sigma})$ \COMMENT{Inverse covariance matrix}
\end{algorithmic}
\end{algorithm}
}

\begin{algorithm}[t]
\caption{Matrix inverse by Newton-Schulz iterations, named $\text{nsInv}(\cdot)$. It is used to compute ``Precision $\mathbf{\Omega}$'' and is also used by Alg. \ref{alg:two}, whereas $\mathbf{Y}_{N_s}$ (line 4) gives iSQRT.}
\label{alg: inv sqrt}
\textbf{Input}: Covariance matrix $\mathbf{\Sigma}$, number of iterations $N_s$.\\
\textbf{Output}: Inverse covariance matrix $\mathbf{\Sigma}^{-1}$.

\begin{algorithmic}[1]
\STATE $\mathbf{Y}_0\!=\!\mathbf{\Sigma}'\! =\!\mathbf{\Sigma}$/trace($\mathbf{\Sigma}$), $\mathbf{Z}_0\!=\!\mIdent$
\FOR{$i=1$ to $N_s$} 
\STATE $\mathbf{P}\!=\!\frac{1}{2}(3\mathbf{I}-\mathbf{Z}_{i-1}\mathbf{Y}_{i-1})$\;
    \STATE $\mathbf{Y}_i = \mathbf{Y}_{i-1}\mathbf{P}\;\;\;\text{and}\;\;\;\mathbf{Z}_i = \mathbf{P} \mathbf{Z}_{i-1}$\;
\ENDFOR
\STATE $\mathbf{Q} = \mathbf{Z}_{N_s}$
\STATE $\mathbf{\Sigma}^{-1} = \mathbf{Q}\mathbf{Q}^\top/\mathrm{trace}(\mathbf{\Sigma})$ \COMMENT{Inverse covariance matrix}
\end{algorithmic}
\end{algorithm}

\vspace{0.1cm}
\noindent\textbf{Estimation of sparse inverse covariance $\mathbf{S}$.} Given the result obtained in the last step, $\mathbf{S}_0$, 
we start iterations of iSICE by applying the projected gradient descend (PGD) to 
the gradient of SICE (Eq. \eqref{eq: sice_obj}), given in Eq. \eqref{eq: isice}.

Following methodology of optimisation by imposing box constraints (\eg, see an intuitive  example by Schiele \etal\cite{box_lasso}), we  separate $\mathbf{S}$ into its positive and negative parts:
%
\begin{equation}
    \mathbf{S}_i^+ =\max(0, \mathbf{S}_{i-1})\quad\text{and}\quad\mathbf{S}_i^- =\max(0, -\mathbf{S}_{i-1}),
\end{equation}
and apply the PGD step to each of them separately.

In such a case, the sparsity  constraint imposed by the $\ell_1$ norm simplifies, \ie, the gradient of $\lambda\mathbf{S}^+$ can be assumed $\lambda$ and the gradient of $-\lambda\mathbf{S}^-$ can be also assumed $\lambda$.
Thus, we firstly rewrite  Eq. \eqref{eq: isice} into two parts:
\begin{equation}
    \nabla\mathbf{S}_i^+ = \mathbf{S}_{i-1}-\mathbf{\Sigma}-\lambda \quad\text{and}\quad-\nabla\mathbf{S}_i^- = \mathbf{S}_{i-1}-\mathbf{\Sigma}+\lambda.
\end{equation}

Then we take one PGD step to update $\mathbf{S}_i^+$ and
$\mathbf{S}_i^-$:
\begin{equation}
    \mathbf{S}_i^+:= \mathrm{\Pi}(  \mathbf{S}_i^+-\eta\beta\nabla\mathbf{S}_{i}^+)\quad\text{and}\quad
    \mathbf{S}_i^-:= \mathrm{\Pi}(\mathbf{S}_{i}^- -\eta\beta\nabla\mathbf{S}_{i}^-),
    \end{equation}
    where  $\Pi(\cdot)\equiv\mathrm{max}(0,\cdot)\equiv\mathrm{ReLU}(\cdot)$ is the gradient reprojection function of PGD into the feasible region of each box constraint (one for the non-negative $\mathbf{S}_i^+$, and one for non-positive $\mathbf{S}_i^-$). 
    Constant $\eta>0$ is a desired learning rate, whereas $\beta>0$ controls the decay of learning rate. 
 
    Finally, 
    we  assemble the current estimate of $\mathbf{S}_i $ from $\mathbf{S}_i^+$ and $\mathbf{S}_i^-$:
  \begin{equation}  
    \mathbf{S}_i = \text{Sym}(\mathbf{S}_i^+ - \mathbf{S}_i^-),
    \label{eq: pgd}
\end{equation}
where $\text{Sym}(\mathbf{M})\!=\!\frac{1}{2}(\mathbf{M}+\mathbf{M}^\top)$ ensures the matrix $\mathbf{M}$ is symmetric (and the intermediate estimate of SICE).

Algorithm \ref{alg:two} starts with a dense precision matrix $\mathbf{S}_0$. If $N>0$, it loops over iterations $i=1,\cdots,N$, applying the above steps. For ease of tuning the learning rate $\eta$, the algorithm starts by the trace normalisation of $\mathbf{\Sigma}$ and it reverses the trace normalisation when it finishes. Otherwise, $\eta$ has to be scaled depending on the value of the largest eigenvalue of $\mathbf{\Sigma}$ which is somewhat impractical when running CNN end-to-end over multiple mini-batches.

\begin{algorithm}[t]
\caption{Iterative sparse inverse covariance estimation (iSICE).}
\label{alg:two}
\textbf{Input}: Sample-based covariance matrix $\mathbf{\Sigma}$, sparsity constant $\lambda$, learning rate $\eta$, number of iterations N, small constant $\alpha$, \ie, $\alpha$ = 1e-9, regularisation parameter $\beta$. \\
\textbf{Output}: Sparse inverse covariance matrix $\mathbf{S}'$.

\renewcommand{\algorithmicthen}{}
\renewcommand{\algorithmicelsif}{\textbf{catch}}
\begin{algorithmic}[1]
\STATE $\nabla_2\!=\!\mathbf{\Sigma}'\!=\!\mathbf{\Sigma}$/trace($\mathbf{\Sigma}$) \COMMENT{Pre-normalisation using trace}$\!\!\!\!$
\STATE $\mathbf{S}_0\!=\!\text{nsInv}(\mathbf{S}_{i-1}\!+\!\alpha\mathbf{I}$) \COMMENT{Fast approx. inverse (Alg. \ref{alg: inv sqrt})}
\STATE $\nabla_1\!=\!\mathbf{S}_0$
\FOR{$i=1$ to $N$} 
\STATE $\mathbf{S}_i^+\!= \mathrm{ReLU}(\mathbf{S}_{i-1})$\quad\text{and}\quad$\mathbf{S}_i^- = \mathrm{ReLU}(-\mathbf{S}_{i-1})$
\IF{$i\neq1$}
\STATE $\nabla_1\!=\!\text{nsInv}(\mathbf{S}_{i-1}\!+\!\alpha\mathbf{I}$) \COMMENT{Fast approx. inv. (Alg. \ref{alg: inv sqrt})}$\!\!\!\!$
\ENDIF
\STATE $\nabla_{12} = \nabla_1-\nabla_2$\;
\STATE $\beta = 1 - \frac{i-1}{\max(1,{N}-1)}$ \COMMENT{Decay the learning rate}
\STATE $\mathbf{S}_i^+\!:= \mathrm{\Pi}\big(\mathbf{S}_i^+\!-\eta\beta(-\nabla_{12} + \lambda)\big)$
\STATE $\mathbf{S}_i^-\!:= \mathrm{\Pi}\big(\mathbf{S}_i^-\!-\eta\beta(+\nabla_{12} + \lambda)\big)$
\STATE $\mathbf{S}_i = \text{Sym}(\mathbf{S}_i^+\!- \mathbf{S}_i^-)$\;
\ENDFOR
\STATE $\mathbf{S}^* = \mathbf{S}_N/{\mathrm{trace}(\mathbf{S}_N)}$
\end{algorithmic}
\end{algorithm}

As  $\mathbf{S}^*$ is a symmetric matrix, we only  take its upper-triangular entries (plus the diagonal entries) and process them by fully connected layers for classification purposes. 

Algorithm \ref{alg:two} is implemented with modern deep learning library, PyTorch, to leverage the full GPU support and autograd package for optimisation. Due to the iterative nature of solving $\mathbf{S}$, we call our method iterative SICE (iSICE).

\section{Experiments}
Below, we first describe experimental dataset benchmarks and then discuss the implementation of our proposed method. Subsequently, we present our experimental results and ablation study on key hyper-parameters. Finally, we compare our proposed method with the existing methods.

\subsection{Datasets, Metric, and Implementation}
\label{exp:data}

\vspace{0.1cm}
\noindent
\textbf{Datasets.} We conduct experiments using one scene  and five fine-grained image datasets: the MIT Indoor dataset \cite{quattoni2009recognizing}, Airplane \cite{maji2013fine}, Birds \cite{wah2011caltech}, Cars \cite{krause20133d}, DTD \cite{cimpoi2014describing} and iNaturalist \cite{van2018inaturalist}. 
We also use ImageNet100 (a subset of ImageNet-1K dataset) proposed by Tian \etal \cite{imagenet100} and mini-ImageNet \cite{vinyals2016matching}.  We follow the widely used training and testing protocols of Bilinear CNN \cite{lin2015bilinear}. The details of datasets and protocols are provided in Appendix \ref{app:data}.

\vspace{0.1cm}
\noindent
\textbf{Metric for  evaluations.} For evaluation of different methods, average classification accuracy is used. This metric is widely used in literature, \eg, \cite{lin2015bilinear, li2018towards}.

\vspace{0.1cm}
\noindent
\textbf{Implementation details.} Our method is implemented using PyTorch 1.9. We use ImageNet-1K pre-trained backbones provided in Torchvision 0.13.0 library. Following the recent works  \cite{li2017second,li2018towards}, the number of feature channels is reduced to 256 with $1\times 1$ convolution for efficiency and fair comparisons. 
All images are resized to $448\times 448$ and the training is conducted by randomly flipping them horizontally. 
We fine-tune all backbones for 50-100 epochs with AdamW optimiser \cite{loshchilov2017decoupled} for an initial learning rate of 0.00012, and ConvNext-T CNN and Swin-T  with an initial learning rate 0.00005. For all backbones, we decrease the learning rate by a factor of 10 at the 15th and 30th epochs. Depending on the dataset and backbones, our fine-tuning process lasts for about 3-8 hours on four P100 GPUs. 
For ImageNet100, ResNet-50 was trained 
for 100 epochs with  the initial learning rate 0.01, reduced by 10 at the 15th, 30th and 45th epochs. 
Settings are detailed in Appendix \ref{app:backb}.

\subsection{Evaluations}
\label{sec:eval}
We  evaluate the performance of the proposed iSICE and compare it with its covariance-based competitors. We also include comparisons with our baseline, the inverse covariance matrix, called precision matrix\footnote{The use of precision matrix (and partial correlations) as a visual representation in place of the sample-based covariance matrix $\mathbf{\Sigma}$ is also our minor but novel proposition. We obtain it via Alg. \ref{alg: inv sqrt} from which we recover the inverse square root $\mathbf{Q}$ and then $\mathbf{\Omega}\!=\!\mathbf{\Sigma}^{-1}\!\!=\!\mathbf{Q}\mathbf{Q}^\top\!/\mathrm{trace}(\mathbf{\Sigma})$.} (for simplicity denoted as Precision $\mathbf{\Omega}$ in experiments). In contrast to iSICE which represents a sparse graph, precision matrix  is used as a baseline as it represents a graph without imposed sparsity. 

 The covariance representations (denoted with COV) are widely used in literature  \cite{lin2015bilinear,lin2017bilinear,li2017second}.  We use the Newton-Schulz iteration\footnote{In contrast to our precision matrix  $\mathbf{\Omega}\!=\!\mathbf{\Sigma}^{-1}\!\!=\!\mathbf{Q}\mathbf{Q}^\top\!/\mathrm{trace}(\mathbf{\Sigma})$ from Alg. \ref{alg: inv sqrt}, iSQRT-COV uses $\mathbf{\Sigma}^{\frac{1}{2}}\!\!=\!\mathbf{Y}_N/\sqrt{\mathrm{trace}(\mathbf{\Sigma})}$ from Alg. \ref{alg: inv sqrt}.} for computing the matrix square root normalised covariance (iSQRT-COV) due to  efficiency of the Newton-Schulz iteration with GPUs, as well as good empirical results reported by multiple authors \cite{li2018towards,lin2017improved}. 

\vspace{0.1cm}
\noindent\textbf{Backbones.} 
We choose VGG-16 \cite{simonyan2014very} and ResNet-50 \cite{he2016deep} CNNs as our backbones for the majority of experiments (we also include the VGG-19, ResNet-101, ResNeXt-101 \cite{xie2017aggregated}, and latest ConvNext-T \cite{liu2022convnet}, Swin-T and Swin-B \cite{liu2021swin} in the main table). VGG-16 and ResNet-50 are popular in 
image classification (including fine-grained benchmarks). We choose these two backbones in order to better understand the performance of our methods compared to baselines in the common testbed (the same backbones and experimental settings). Given an input image, we obtain a set of feature channels shaped as a tensor after the $1\times 1$ convolution operation. Using these feature channels, we compute the iSICE, iSQRT-COV and Precision $\mathbf{\Omega}$ representations. Since these representations are symmetric, we only use the upper-triangular entries (and the diagonal entries) passed to a fully-connected layer to obtain classification scores.

\vspace{0.1cm}
\noindent\textbf{Hyper-parameters.} There are three hyper-parameters associated with iSICE: sparsity constant $\lambda$, learning rate $\eta$ and number of iterations $N$. We experiment with a large range of values for a better understanding, \ie, $\lambda\!\in$\{1.0, 0.5, 0.1, \textbf{0.01}, 0.001, 0.0001, 0.00001\}, $\eta\!\in$\{0.001, 0.01, 0.1, \textbf{1.0}, 5.0, 10.0, 20.0\}, and $N\!\in$\{1, \textbf{5}, 10\}. Since the total combination of hyper-parameters in the table is 147, we  choose the median values of each hyper-parameter range (highlighted in bold) and keep them throughout experiments on all datasets. 
Appendix \ref{app:hyper} studies the impact of hyper-parameters on results.

\begin{table*}[t]
\vspace{-0.3cm}
\setlength{\tabcolsep}{20pt}
\centering

\resizebox{1\linewidth}{!}{
\begin{tabular}{lccccccccc}
\toprule
Method & Backbone & MIT & Airplane & Birds & Cars & DTD & iNatuarlist & mini-ImageNet \\
\midrule
GAP \cite{simonyan2014very} & \multirow{19}{*}{VGG-16} & -- & 76.6 & 70.4 & 79.8 & -- & -- & --\\
NetVLAD \cite{arandjelovic2016netvlad} && -- & 81.8 & 81.6 & 88.6 & -- & -- & --  \\
NetFV \cite{lin2017bilinear} && -- & 79.0  & 79.9 & 86.2 & -- & -- & -- \\
BCNN \cite{lin2015bilinear} && 77.6 & 83.9 & 84.0 & 90.6 & 84.0 & -- & -- \\
CBP \cite{gao2016compact} && 76.2 & 84.1 & 84.3 & 91.2 & 84.0 & -- & --  \\
LRBP \cite{kong2017low} && -- & 87.3 & 84.2 & 90.9 & -- & -- & --  \\
KP \cite{cui2017kernel} && -- & 86.9 & 86.2 & 92.4 & -- & -- & --   \\
HIHCA \cite{cai2017higher} && -- & 88.3 & 85.3 & 91.7 & -- & -- & -- \\
Improved BCNN \cite{lin2017improved} && -- & 88.5 & 85.8 & 92.0 & -- & -- & -- \\
SMSO \cite{yu2018statistically} && 79.5 & -- & 85.0 & -- & -- & -- & -- \\
MPN-COV \cite{wang2020deep} (reproduced) && -- & 86.1 & 82.9 & 89.8 & -- & -- & --  \\
iSQRT-COV \cite{li2018towards} (reproduced) && 76.1 & 90.0 & 84.5 & 91.2 & 71.3 & 56.2  & 76.2 \\
DeepCOV \cite{engin2018deepkspd} && 79.2 & 88.7 & 85.4 & 91.7 & 86.3 & -- & --  \\
DeepKSPD \cite{engin2018deepkspd} && \textbf{81.0} & 90.0 & 84.8 & 91.6 & 86.3 & -- & -- \\
RUN \cite{yu2020toward} && 80.5 & 91.0 & 85.7 & -- & -- & -- & --  \\
FCBN \cite{yu2021fast} && 80.3 & 90.5 & 85.5 & -- & -- & -- & --  \\
TKPF \cite{yu2022fast} && 80.5 & 91.4 & 86.0 & -- & -- & -- & --  \\ 

Precision $\mathbf{\Omega}$  && 80.2 & 89.4 & 83.4 & 92.0 & 74.0 & 57.9 & 74.0  \\
\rowcolor{LightCyan}  iSICE (ours)&& 78.7 & \textbf{92.2} & \textbf{86.5} & \textbf{94.0} & \textbf{74.7} & \textbf{59.8} & \textbf{78.7} \\ \midrule

CBP \cite{gao2016compact} & \multirow{9}{*}{ResNet-50} & -- & 81.6 & 81.6 & 88.6 & -- & -- & --  \\
KP \cite{cui2017kernel} && -- & 85.7 & 84.7 & 91.1 & -- & -- & --   \\
SMSO \cite{yu2018statistically} && 79.7 & -- & 85.8 & -- & -- & -- & --   \\
iSQRT-COV \cite{li2018towards} (reproduced) && 78.8 & 90.9 & 84.3 & 92.1 & 73.0 & 57.7 & 70.7   \\
DeepCOV-ResNet \cite{rahman2020redro} && 83.4 & 83.9 & \textbf{86.0} & 85.0 & 84.6 & -- & -- \\ 
TKPF \cite{yu2022fast} && \textbf{84.1} & 92.1 & 85.7 & -- & -- & -- & --  \\ 

Precision $\mathbf{\Omega}$  && 80.8 & 91.2 & 84.7 & 92.0 & 73.7 & 59.6 & 65.6   \\
\rowcolor{LightCyan} iSICE (ours)&& 80.5 & \textbf{92.7} & 85.9 & \textbf{93.5} & \textbf{60.7} & \textbf{60.7} & \textbf{72.0}  \\ \midrule

iSQRT-COV \cite{li2018towards} & \multirow{3}{*}{VGG-19} & 76.3 & 90.3 & 84.1 & 91.4 & 71.8 & 56.9 & 75.4  \\
Precision $\mathbf{\Omega}$ && 79.6 & 91.1    & 83.2  & 92.2 & 74.2 & 57.3 & 73.8  \\
\rowcolor{LightCyan} iSICE (ours) && \textbf{80.6} & \textbf{92.5} & \textbf{86.6} & \textbf{93.9} & \textbf{74.9} & \textbf{59.6} & \textbf{77.1} \\ \midrule

iSQRT-COV \cite{li2018towards} & \multirow{3}{*}{ResNet-101} & 79.3 & 91.0    & 84.4  & 92.3 & 73.0 & 70.6 & 73.9 \\
Precision $\mathbf{\Omega}$ && 77.9 & 90.1    & 83.3  & 91.4 & 71.2 & 69.8 & 73.0\\
\rowcolor{LightCyan} iSICE (ours) && \textbf{81.0} & \textbf{92.9}    & \textbf{86.6}  & \textbf{93.6} & \textbf{75.4} & \textbf{72.0} & \textbf{78.0} \\ \midrule

iSQRT-COV \cite{li2018towards} & \multirow{3}{*}{ResNeXt-101} & 81.6 & 91.3    & 86.2  & 92.4 & 75.7 & 72.2 & 76.1 \\
Precision $\mathbf{\Omega}$ && 85.7 & 90.2    & 84.6  & 89.9 & 76.9 & 72.3 & 77.6  \\
\rowcolor{LightCyan} iSICE (ours) && \textbf{86.3} & \textbf{94.6}    & \textbf{87.2}  & \textbf{94.5} & \textbf{78.7} & \textbf{73.8} & \textbf{81.0}\\ \midrule

iSQRT-COV \cite{li2018towards} & \multirow{3}{*}{ConvNext-T} & 77.8 & 88.1 & 83.5 & 89.4 & 84.7 & 61.5 & 82.0\\
Precision $\mathbf{\Omega}$ && 78.5 & 81.2 & 83.7 & 92.2 & 83.9 & 59.3 & 83.6 \\
\rowcolor{LightCyan} iSICE (ours)&& \textbf{85.4} & \textbf{90.4} & \textbf{86.7} & \textbf{93.1} & \textbf{88.9} & \textbf{65.0} & \textbf{85.1}\\ \midrule

iSQRT-COV \cite{li2018towards} & \multirow{3}{*}{Swin-T} & 82.1 & 87.6 & 85.1 & 89.7 & 86.1 & 58.1 & 67.7 \\
Precision $\mathbf{\Omega}$  & & 82.5 & 88.2 & 84.9 & 90.5 & 86.5 & 59.1 & 65.6 \\
\rowcolor{LightCyan}  iSICE (ours)&& \textbf{85.9} & \textbf{89.6} & \textbf{86.5} & \textbf{91.3} & \textbf{88.3} & \textbf{61.9} & \textbf{69.1} \\ \midrule

iSQRT-COV \cite{li2018towards} & \multirow{3}{*}{Swin-B} & 86.6 & 91.3    & 88.0  & 92.0 & 79.4 & 69.7 & 64.9  \\
Precision $\mathbf{\Omega}$  & & 87.0 & 90.7    & 87.7  & 93.1 & \textbf{80.1} & 67.3 & 66.4 \\
\rowcolor{LightCyan}  iSICE (ours)&& \textbf{87.6} & \textbf{92.9}    & \textbf{88.3}  & \textbf{93.3} & 79.8 & \textbf{72.4} & \textbf{68.4} \\
\bottomrule
\end{tabular}}
\caption{Comparison between iSICE, Precision $\mathbf{\Omega}$ and other SPD representations in terms of classification accuracy (\%). The performance of existing SPD representation methods is quoted from the original papers. Precision $\mathbf{\Omega}$ is given by Alg. \ref{alg: inv sqrt}. iSICE is given by Alg. \ref{alg:two}.}
\label{tab: comparison-with-all}
\end{table*}

\begin{table}[t]
    \centering
    \setlength{\tabcolsep}{18pt}
    \resizebox{1\linewidth}{!}{
    \begin{tabular}{lccc}
    \toprule
    Method &  Backbone & Top-1 & Top-5\\
    \midrule
    GAP \cite{he2016deep} & \multirow{4}{*}{\makecell{ResNet-50/\\VGG-16}} & 71.0/69.5 & 90.9/88.9 \\
    iSQRT-COV \cite{li2018towards} && 71.5/70.2 & 90.5/89.7\\
    Precision $\mathbf{\Omega}$ && 71.1/71.0 & 90.1/90.1\\
    \rowcolor{LightCyan} iSICE && \textbf{74.8}/\textbf{73.4} & \textbf{92.0}/\textbf{91.8} \\
    \bottomrule
    \end{tabular}}
    \caption{Results on the ImageNet100 dataset.}
    \label{tab:in100}
    \vspace{-0.3cm}
\end{table}

\vspace{0.1cm}
\noindent\textbf{Overview of results.} Table 
\ref{tab: comparison-with-all} shows the performance of several COV models (\eg, popular iSQRT-COV), and our Precision $\mathbf{\Omega}$ and iSICE models. 
The rightmost column summarizes the average performance over one scene and three fine-grained image classification benchmarks.  
It is clear that on average, iSICE outperforms MPN-COV, iSQRT-COV, DeepCOV, and DeepKSPD, \etc. iSICE also outperforms our baseline Precision $\mathbf{\Omega}$. This achievement is consistent in all four backbones. It is interesting to see that except a few cases, the inverse covariance method, \ie, Precision $\mathbf{\Omega}$, performs slightly better than the covariance method, \ie, iSQRT-COV. This improved performance highlights the effectiveness of characterising partial correlations of features with inverse covariance instead of pairwise correlations of features based on the sample covariance. Our iSICE method makes the inverse covariance estimation more robust and reliable by enforcing sparsity, as demonstrated by improved performance over Precision $\mathbf{\Omega}$ baseline. However, there may be some situations when Precision $\mathbf{\Omega}$  outperforms iSICE (\eg, MIT with VGG-16 backbone). Notice that iSICE can be considered as sparse precision matrix. When $N=0$ in Alg. \ref{alg:two}, iSICE reduces to Precision $\mathbf{\Omega}$. In further experiments we show that once the size of matrix is increased, iSICE does outperform Precision $\mathbf{\Omega}$. 

Table \ref{tab:in100} corroborates that iSICE significantly outperforms  Precision $\mathbf{\Omega}$ and iSQRT-COV.

\noindent\textbf{Robustness of iSICE to Hyper-parameters.} 
We have conducted  experiments with the hyper-parameter range given in Section \ref{exp:data}. 
 Appendix \ref{app:hyper} shows that the performance of iSICE remains stable across a range of values.

\begin{table*}[t]
\vspace{-0.3cm}
\setlength{\tabcolsep}{16pt}
\centering
\resizebox{1\linewidth}{!}{
\begin{tabular}{llllllllllll}
\toprule
\multirow{2}{*}{Method} & \multirow{2}{*}{Matrix Dim.} & \multicolumn{2}{c}{MIT} & \multicolumn{2}{c}{Airplane} & \multicolumn{2}{c}{Birds} & \multicolumn{2}{c}{Cars} & \multicolumn{2}{c}{Average} \\ \cline{3-12}
&  & VGG & ResNet & VGG & ResNet & VGG & ResNet & VGG & ResNet & VGG & ResNet \\ \midrule
\multirow{2}{*}{iSQRT-COV} & $256\times256$ & 76.1 & 78.8 & 90.0   & 90.9 & 84.5 & 84.3 & 91.2 & 92.1 & 85.5 & 86.5 \\
\rowcolor{LightCyan}\cellcolor{white}\multirow{-2}{*}{iSQRT-COV}& $512\times 512$ & 76.9 & 82.8 & 91.5 & 91.1 & 85.0  & 84.5 & 92.2 & 92.1 & 86.4 & 87.6 \\ \midrule
\multirow{2}{*}{Precision $\mathbf{\Omega}$} & $256\times256$ & 80.2 & 80.8 & 89.4 & 91.2 & 83.4 & 84.7 & 92.0   & 92.0   & 86.3 & 87.1 \\
\rowcolor{LightCyan}\cellcolor{white}\multirow{-2}{*}{Precision $\mathbf{\Omega}$} & $512\times 512$ & 80.7 & 82.7 & 90.1 & 91.5 & 84.9 & 84.0   & 92.5 & 92.6 & 87.0   & 87.7 \\ \midrule

\multirow{2}{*}{SICE} & $128\times128$ & 71.0 & 73.1 & 85.5 & 86.9 & 77.3 & 78.0 & 87.0 & 87.9 & 80.2 & 81.5\\

\rowcolor{LightCyan}\cellcolor{white}\multirow{-2}{*}{SICE} & $256\times256$ & 73.7 & 75.4 & 87.9 & 89.2 & 79.7 &  80.3 & 89.5 & 89.3 & 82.7 & 83.6\\ \midrule

\multirow{2}{*}{iSICE} & $256\times256$ & 78.7 & 80.5 & 92.2 & 92.7 & 86.5 & 85.9 & 94   & 93.5 & 87.9 & 88.2 \\
\rowcolor{LightCyan}\cellcolor{white}\multirow{-2}{*}{iSICE} & $512\times 512$ & 81.1 & 81.7 & 92.9 & 92.6 & 86.8 & 86   & 94.6 & 93.8 & 88.9 & 88.5 \\
\bottomrule
\end{tabular}
}
\caption{Performance of iSQRT-COV, Precision $\mathbf{\Omega}$, SICE and iSICE on various datasets when different matrix dimensions  are used. 
}
\label{tab: sice size}
\vspace{-0.3cm}
\end{table*}

\vspace{0.1cm}
\noindent\textbf{Detailed comparisons with  SPD-based SOTA models.} 
Table \ref{tab: comparison-with-all} compares the performance of iSICE to several prior works. 
We first compare our VGG-16 backbone based iSICE with BCNN, CBP, LRBP, KP, HIHCA, Improved BCNN, SMSO, MPN-COV, iSQRT-COV, DeepCOV, DeepKSPD, RUN, FCBN and TKPF methods. The MPN-COV and iSQRT-COV methods use backbones pre-trained with second-order pooling. For fair comparison with iSICE, we re-run those methods on our machine with the same backbone and evaluation protocols as ours. iSICE outperforms all existing methods on fine-grained datasets. On MIT dataset, our performance is better than CBP, BCNN and iSQRT-COV. iSICE could outperform DeepKSPD and other methods on MIT if a large-dimensional matrix similar to those is used (see Table \ref{tab: sice size}). 

Secondly, we compare our ResNet-50 backbone based iSICE with CBP, KP, SMO, iSQRT-COV, DeepCOV-ResNet and TKPF methods. iSICE achieves better performance than existing methods on both Airplane and Cars datasets. DeepCOV-ResNet uses $1024\times 1024$-dimensional matrix which is four times larger than ours. TKPF uses an advanced feature projection to reduce the CNN feature channels and we use a simple linear projection with $1\times 1$ convolution. However, on average, we are still better than DeepCOV-ResNet and TKPF.

Thirdly, we integrate iSICE with the popular VGG-19, ResNet-101, ResNeXt-101, ConvNext-T \cite{liu2022convnet}, Swin-T and Swin-B \cite{liu2021swin} backbones pre-trained on ImageNet-1K, and compare their performance with iSQRT-COV and Precision $\mathbf{\Omega}$ (Alg. \ref{alg: inv sqrt}). 
iSICE outperforms  iSQRT-COV and Precision $\mathbf{\Omega}$ methods across all datasets. 
This highlights that SPD-based visual representations (1) are still relevant for modern powerful classification backbones and (2) they improve results on large-scale datasets.

\vspace{0.1cm}
\noindent\textbf{Ablations on the size of Sparse Inverse Matrix.} 
Table \ref{tab: sice size} shows results  for $512\times 512$ \vs $256\times 256$ matrix size (we keep the hyper-parameters fixed). 
To produce $512\times 512$ dim. matrix, (1) from VGG-16, we simply remove the $1\times 1$ convolution layer to obtain 512 feature channels and (2) from ResNet-50, we set the output channels of $1\times 1$ convolution layer to 512. Generally, switching to a larger matrix improves the performance, \eg, on  MIT  our iSICE gains between 1.2 and 2.4\% (sparsity helps with a larger matrix). 
On average, all methods improved performance  by switching to a larger matrix at the  computational expense. See  runtimes in Table \ref{tab:runt}  and memory consumption in Appendix \ref{app:memory comsumption}. Finally, Table \ref{tab: sice size} also shows that iSICE performs much better than SICE  (based on ADMM solver \cite{boyd2011distributed}) computed over pre-extracted features. Table \ref{tab:runt} shows that iSICE is $3\times$ faster. SICE is almost intractable on larger datasets. This validates our claim that learning sparse inverse covariance matrix 
end-to-end produces robust visual representation.

\vspace{0.1cm}
\noindent
\textbf{iSICE with learning rate and sparsity modulators.} As iSICE 
trades between $\log\det(\cdot)$ (changes rapidly) and the $\ell_1$ norm (changes linearly), optimizing Eq. \eqref{eq: sice_obj} with PGD may struggle with non-optimal learning rates and sparsity. Thus, we design a simple modulator that updates $\beta$ in Alg. \ref{alg:two} by setting $\beta:=\beta\!\cdot\!\kappa$ in lines 10, where 
$\kappa\!=\!\tau\!+\!2\text{Sigm}\big(\text{FC}(\mathbf{X}\mathbf{1}/n) \big)$,  
 $\text{Sigm}(\cdot)$ is a sigmoid, and FC layer is  of $d\times1$ size. 
We also add a penalty $-\gamma(\kappa\!-\!1)^2$ to the classification loss to encourage $\kappa$ to be close to 1 unless classification loss gets smaller for $\kappa\!\neq\!1$ while incurring the above penalty. We set $\gamma\!=\!0.0001$. We use the above modulator (we do not claim this is the most optimal design) as a tool akin to ModGrad \cite{christian_modgrad}. $\tau\!=\!0.01$ is a small offset to prevent zero learning rate. Another modulator with the same architecture is used to adapt sparsity parameter $\lambda$.  Table \ref{tab:beta-exp} shows that modulating the learning rate and sparsity on-the-fly helps iSICE.

 \begin{table}[htbp]
    \centering
    \setlength{\tabcolsep}{12pt}
    \resizebox{1\linewidth}{!}{
    \begin{tabular}{lccccc}
    \toprule
     Method &  MIT & Airplane & Birds & Cars & ImageNet100\\ \midrule
     iSICE & 80.5 & 92.7 & 85.9 & 93.5 & 74.8\\
     \rowcolor{LightCyan}  iSCIE+MLP & 81.3 & 93.4 & 86.1 & 93.9 & 76.3\\
     \bottomrule
    \end{tabular}
    }
    \caption{Comparison between the classification performance of iSICE and iSICE+MLP on the ResNet-50 backbone.}
    \label{tab:beta-exp}
\end{table}

\begin{table}[]
    \centering
    \resizebox{1\linewidth}{!}{
    \begin{tabular}{lcccccc}
    \toprule
         &  GAP & iSQRT-COV & Precision $\Omega$ & SICE & iSICE & iSICE+MLP \\
         \midrule
     Time/batch (sec.)& 29.0 & 32.0 & 32.8 & 150.8 & 44.6 & 45.8\\ \midrule
     Time/epoch (min.)& 12.6 & 13.4 & 13.8 & 65.3 & 19.3 & 19.8\\
     \bottomrule
     
    \end{tabular}
    }
    \caption{Runtimes ($256\!\times\!256$ matrix,  ImageNet100, ResNet-50).}
    \label{tab:runt}
    \vspace{-0.2cm}
\end{table}

\begin{figure}
    \centering
    \includegraphics[width=1\linewidth]{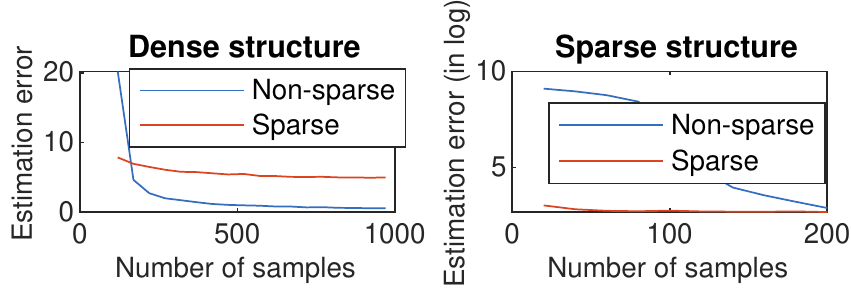}
     \vspace{-0.2cm}
    \caption{Estimation errors on (left) dense \& (right) sparse struct.}
    \label{fig:my_label2}
    \vspace{-0.2cm}
\end{figure}

\vspace{0.1cm}
\noindent\textbf{Experiments on dense \vs sparse structure estimation \wrt sample size.} Below we randomly generate a dense 
or sparse inverse covariance matrix ${\mathbf P}$ (size $100\!\times\!100$) and sample various amount of data from the resulted normal distribution (via mvnrnd($\cdot$) in Matlab) to get its estimate $\hat{\mathbf P}$. Fig. \ref{fig:my_label2} (left) shows estimation error $\|\hat{\mathbf P}-{\mathbf P}\|_{F}$ for dense structure. Sparse and non-sparse estimation (by SICE and MLE, resp.) show high errors in low sample regime. 
For the sparse structure in Fig. \ref{fig:my_label2} (right), sparse estimation works better. 

\section{Conclusions}
In this paper, we proposed a method for learning sparse inverse covariance representation with CNN. Our method estimates SICE within the CNN layers and facilitates backpropagation for end-to-end training. Our iSICE significantly outperforms 
other covariance representations on several datasets. iSICE 
exploits the sparsity prior to capture partial correlations under limited number of samples.  Our method is of general purpose and can be readily applied in existing SPD-based models to improve their performance. 

\vspace{0.1cm}
\noindent\textbf{Acknowledgements.} This work was supported by CSIRO’s Machine Learning \& Artificial Intelligence Future Science Platform (MLAI FSP), University of Wollongong's Australia IPTA scholarship, and Australian Research Council (DP200101289). We thank Jianjia Zhang for discussions.

{\small
\bibliographystyle{ieee_fullname}
\bibliography{references}
}

\clearpage
\appendix

\title{Learning Partial Correlation based Deep Visual Representation for Image Classification (Supplementary Material)}

\author{Saimunur Rahman$^{1,2}$, Piotr Koniusz$^{*,1,3}$, Lei Wang$^2$, Luping Zhou$^4$, Peyman Moghadam$^{1,5}$, Changming Sun$^1$\\
$^1$Data61\heart CSIRO, $^2$University of Wollongong, $^3$Australian National University, \\ $^4$University of Sydney $^5$Queensland University of Technology \\
{\tt\small name.surname@data61.csiro.au, leiw@uow.edu.au, luping.zhou@sydney.edu.au}
}
\maketitle

\section{Datasets and the Evaluation Protocols}
\label{app:data}
In this section, we provide the details of datasets and their evaluation protocols (see Section \ref{exp:data} of the main text). We perform experiments on eight widely used public image datasets, namely, MIT Indoor \citelatex{quattoni2009recognizing_supp}, Stanford Cars \citelatex{krause20133d_supp}, Caltech-UCSD Birds (CUB 200-2011) \citelatex{wah2011caltech_supp}, FGVC-Aircraft \citelatex{maji2013fine_supp},  DTD \cite{cimpoi2014describing}, iNaturalist \cite{van2018inaturalist}, mini-ImageNet \cite{vinyals2016matching} and ImageNet100 \citelatex{imagenet100_supp} to demonstrate the performance of our methods. Figure \ref{fig:datasets-sample-images} shows the sample images from the datasets. Further details are given below.

\begin{figure*}[t]
\centering
\includegraphics[width=1\textwidth]{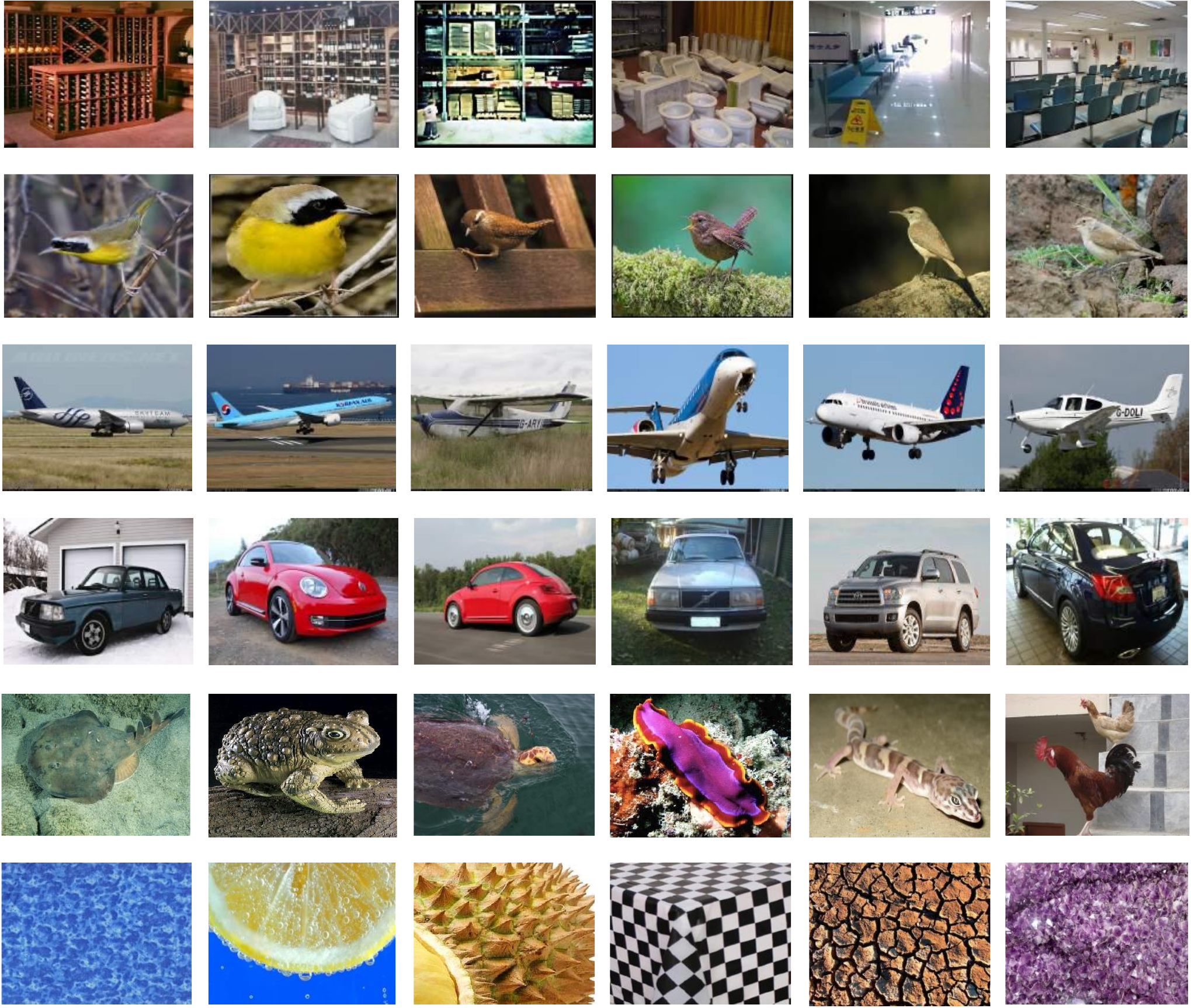}
\caption{Sample images from the datasets used in our experiments. Rows 1, 2, 3, 4, 5 and 6 have the images from MIT Indoor, Caltech-UCSD Birds, FGVC-Aircraft, Stanford Cars, ImageNet100/mini-ImageNet and DTD datasets, respectively.}
\label{fig:datasets-sample-images}
\end{figure*}

\vspace{0.1cm}
\noindent
\textbf{MIT Indoor} dataset is one of the most widely used datasets in the literature for scene classification. It has a total of 15,620 images and 67 classes. Each image class contains a minimum number of 100 images. The images are collected from various types of stores (\eg, grocery, bakery), private places (\eg, bedroom and living room), public places (\eg, prison cell, bus, library), recreational places (\eg, restaurant, bar) and working environments (\eg, office, studio).

\vspace{0.1cm}
\noindent
\textbf{Caltech-UCSD Birds} or simply `Birds' is one of the most reported datasets in fine-grained image classification (FGIC) literature. It has a total of 11,788 images and 200 image classes. There are subtle differences between these classes and they are hard to be distinguished by human observers. This dataset comes with bounding box annotations; however, we do not use any annotations in our experiments.

\vspace{0.1cm}
\noindent
\textbf{FGVC-Aircraft} or `Aircraft' dataset is  widely used by many recent FGIC methods. It has only 10,000 images distributed among 100 aircraft classes, and each class has precisely 100 images. Similar to Birds, the classes have subtle differences between them and are hard for humans to distinguish from each other.

\vspace{0.1cm}
\noindent
\textbf{Stanford Cars} or simply `Cars' has a total of 16,185 images and 196 classes. The classes are organized as per the car production year, car manufacturer and car model. Cars dataset has relatively smaller objects, \ie, cars, than those of the airplane dataset. Furthermore, the objects are appeared in cluttered backgrounds.

\vspace{0.1cm}
\noindent
\textbf{Describable Texture Dataset} (DTD) has a total of 5,640 images and 47 classes. The images in all classes represents about 95\% of their class attributes. For evaluation, it has 10 splits and each split has an equal number of images from each class for training, validation and test sets. The average performance across all splits is reported. We used both training and validation sets for training.

\vspace{0.1cm}
\noindent
\textbf{iNaturalist} is a large fine-grained dataset. It has a total of 675,170 images and 5,089 classes. The classes are from 12 super-classes. The dataset is challenging due to its high class-imbalance. We use the training strategy and train-test splits specified in the original paper \cite{van2018inaturalist}.

\vspace{0.1cm}
\noindent
\textbf{mini-ImageNet} is a subset of ImageNet-1K dataset first proposed by Ravi \etal \citelatex{ravi2017optimization}. It contains a total of 60,000 images and 100 classes. We use the original  224$\times$224 image resolution on this dataset. Since the dataset was originally proposed by the authors for few-shot learning tasks, we divide the dataset in a 90:10 ratio for training and testing. We report the average accuracy of 5 runs.

\vspace{0.1cm}
\noindent
\textbf{ImageNet100} is a subset of ImageNet-1K Dataset from ImageNet Large Scale Visual Recognition Challenge 2012. It contains random 100 classes proposed by Tian \etal \citelatex{imagenet100_supp}. ImageNet100 train and validation sets contain 1300 and 50 images per class, respectively. We use the original  224$\times$224 image resolution on this dataset.

\begin{table*}[t]
\centering
\begin{tabular}{lccccc}
\toprule
\multirow{2}{*}{Dataset} & \multirow{2}{*}{\makecell[l]{Total classes}} & \multirow{2}{*}{\makecell[l]{Total images}} & \multicolumn{2}{c}{Predefined protocol} & \multirow{2}{*}{\makecell[l]{Major difficulty}}\\ \cmidrule{4-5}
 &  &  & \makecell[l]{Training images} & \makecell[l]{Testing images} & \\
\midrule
MIT Indoor          & 67    & 6,700      & 5,360     & 1,340 & difficult environment  \\ 
Birds  & 200   & 11,788    & 5,994     & 5,794 & subtle class difference \\ 
Aircraft       & 100   & 10,000    & 6,600     & 3,400 & subtle class difference\\
Cars       & 196   & 16,185    & 8,144     & 8,041 & cluttered background\\
DTD         & 47 & 5,640 & 4512 & 1128 & complex structure \\
iNaturalist & 5,089 & 675,170 & 579,184 & 95,986 & class imbalance \\
mini-ImageNet & 100 & 60,000 & 54,000 & 6,000 & difficult environment\\
ImageNet100 & 100 & 1,35,000 & 130,000 & 5,000 & difficult environment\\
\bottomrule
\end{tabular}
\caption{Summary of datasets.}
\label{table:dataset-summary}
\end{table*}

Table \ref{table:dataset-summary} gives a more concise summary of the datasets. We train our method using the respective training splits provided by the original authors of the datasets. For evaluation, we again use the respective original test splits. This also applies to all of the methods we have compared in the main text. During training, for all datasets except iNaturalist, we resize our images to $448\times 448$ following the work of \citelatex{lin2015bilinear_supp,li2018towards_supp} and use only horizontal flipping as a data augmentation. For iNaturalist, as in original paper, we resize images to $299\times 299$.

\begin{table*}[]
  \centering
  \resizebox{\textwidth}{!}{\begin{tabular}{lccccccccccc}
    \toprule
    &&&& \multicolumn{8}{c}{iSICE} \\  \cmidrule{5-12}
    \multirow{1}{*}{Dataset} & \multirow{1}{*}{Backbone} & \multirow{1}{*}{iSQRT-COV \cite{li2018towards}} & \multirow{1}{*}{Precision $\mathbf{\Omega}$} & \multicolumn{7}{c}{Sparsity constant $\lambda$} & \multirow{2}{*}{\makecell{Mean $\pm$ Std.\\(iSICE)}}\\
    \cmidrule{5-11}
    & & & & 1.0 & 0.5 & 0.1 & 0.01 & 0.001 & 0.0001 & 0.00001  \\
    \midrule
    \multirow{2}{*}{MIT} & VGG-16 & 76.12 & \textbf{80.15} & 77.46 & 78.13 & 78.13 & 78.66 & 78.58 & 78.96 & 78.96 & 78.41$\pm$0.54 \\\cmidrule{2-12}
                         & ResNet-50 & 78.81 & 80.75 & 78.43 & 80.75 & 80.45 & 80.52 & 80.90 & 80.37 & \textbf{81.34} & 80.39$\pm$0.93\\ \cmidrule{1-12}
    \multirow{2}{*}{Airplane} & VGG-16 & 90.01 & 89.44 & 92.26 & 92.71 & 92.77 & 92.23 & \textbf{92.83} & 92.74 & 92.44 & 92.56 $\pm$0.25\\\cmidrule{2-12}
                         & ResNet-50 & 90.88 & 91.15 & \textbf{92.89} & 92.65 & \textbf{92.89} & 92.74 & 92.83 & 92.56 & 92.56 & 92.73$\pm$0.14 \\ \cmidrule{1-12}
    \multirow{2}{*}{Birds} & VGG-16 & 84.47 & 83.36 & 86.04 & 86.47 & 86.35 & \textbf{86.52} & 85.59 & 86.31 & 86.28 & 86.22$\pm$0.32\\\cmidrule{2-12}
                         & ResNet-50 & 84.26 & 84.67 & 84.62 & 85.16 & 85.30 & 85.90 & \textbf{86.05} & 85.90 & 85.59 & 85.50$\pm$0.51\\ \cmidrule{1-12}
    \multirow{2}{*}{Cars} & VGG-16 & 91.21 & 92.04 & 93.60 & 93.98 & \textbf{94.06} & 94.03 & 93.88 & 93.91 & 93.50 & 93.85$\pm$0.22\\ \cmidrule{2-12}
                         & ResNet-50 & 92.13  & 91.99 & 93.01 & 93.36 & 93.69 & 93.51 & 93.22 & \textbf{93.72} & 93.40 & 93.41$\pm$0.25\\
    \bottomrule
  \end{tabular}}
  \caption{Performance of iSICE on changing the sparsity constant $\lambda$ while fixing the learning rate $\eta$ and the number of iterations ${N}$ to 1.0 and 5, respectively. The mean and standard deviation (std.) of the classification performance resulted by iSICE are also shown for a better understanding of sparsity constant changes in iSICE. Results are shown on multiple datasets with VGG-16 and ResNet-50 backbones. The best results in each row are highlighted with boldface.}
  \label{tab: sparsity change}
\end{table*}

\begin{table*}[]
  \centering
  \resizebox{\textwidth}{!}{\begin{tabular}{lccccccccccc}
    \toprule
    &&&& \multicolumn{8}{c}{iSICE} \\  \cmidrule{5-12}
    \multirow{1}{*}{Dataset} & \multirow{1}{*}{Backbone} & \multirow{1}{*}{iSQRT-COV \cite{li2018towards}} & \multirow{1}{*}{Precision $\mathbf{\Omega}$} & \multicolumn{7}{c}{Learning rate $\eta$} & \multirow{2}{*}{\makecell{Mean $\pm$ Std.\\(iSICE)}}\\
    \cmidrule{5-11}
    & & & & 0.001 & 0.01 & 0.1 & 1.0 & 5.0 & 10.0 & 20.0  \\
    \midrule
    \multirow{2}{*}{MIT} & VGG-16 & 76.12 & 80.15 & 78.81 & 79.33 & 77.91 & 78.66 & 78.28 & \textbf{80.52} & 77.76 & 78.75$\pm$0.95\\ \cmidrule{2-12}
                         & ResNet-50 & 78.81 & 80.75 & 80.82 & 80.82 & 80.75 & 80.52 & \textbf{81.19} & 79.33 & 78.96 & 80.34$\pm$0.85\\ \midrule
    \multirow{2}{*}{Airplane} & VGG-16 & 90.01 & 89.44 & 92.32 & 92.38 & 92.98 & 92.23 & \textbf{93.28} & 92.50 & 92.26 & 92.56$\pm$0.41\\\cmidrule{2-12}
                         & ResNet-50 & 90.88 & 91.15 & 92.77 & \textbf{93.01} & 92.89 & 92.74 & 92.65 & 92.95 & 92.38 & 92.77$\pm$0.21\\ \midrule
    \multirow{2}{*}{Birds} & VGG-16 & 84.47 & 83.36 & 86.54 & 86.31 & 86.40 & 86.52 & \textbf{86.73} & 86.59 & 86.45 & 86.51$\pm$0.14\\\cmidrule{2-12}
                         & ResNet-50 & 84.26 & 84.67 & 85.69 & 85.88 & 85.81 & 85.90 & 85.59 & 85.71 & \textbf{85.97} & 85.79$\pm$0.14\\ \midrule
    \multirow{2}{*}{Cars} & VGG-16 & 91.21 & 92.04 & 93.83 & 93.65 & 93.69 & \textbf{94.03} & 93.66 & 93.82 & 93.93 & 93.80$\pm$0.14\\\cmidrule{2-12}
                         & ResNet-50 & 92.13 & 91.99 & \textbf{93.60} & 93.57 & 93.32 & 93.51 & 93.57 & 93.35 & \textbf{93.60} & 93.50$\pm$0.12\\
    \bottomrule
  \end{tabular}}
  \caption{Performance of iSICE on changing the learning rate $\eta$ while fixing the sparsity constant $\lambda$ and the number of iterations ${N}$ to 0.01 and 5, respectively. The mean and standard deviation (std.) of the classification performance resulted by iSICE are also shown for a better understanding of learning rate changes in iSICE. Results are shown on multiple datasets with VGG-16 and ResNet-50 backbones. The best results in each row are highlighted with boldface.}
  \label{tab: lr change}
\end{table*}

\begin{table*}[t]
  \resizebox{\textwidth}{!}{\begin{tabular}{lccccccccc}
    \toprule
    \multirow{2}{*}{Method} & \multirow{2}{*}{Iter. ${N}$} & \multicolumn{4}{c}{Based on VGG-16 backbone} & \multicolumn{4}{c}{Based on ResNet-50 backbone} \\ \cmidrule{3-6} \cmidrule{7-10}
    & & MIT & Airplane & Birds & Cars & MIT & Airplane & Birds & Cars\\
    \midrule
    iSQRT-COV \cite{li2018towards} & 5 &76.12 & 90.01 & 84.47 & 91.21	& 78.81 & 90.88	&84.26 & 92.13 \\
    Precision $\mathbf{\Omega}$ & 7 & \textbf{80.15} & 89.44 & 83.36 & 92.04 & 80.75 & 91.15 & 84.67 & 91.99 \\ 
    \midrule \midrule
    \multirow{3}{*}{iSICE}& 2 & 78.28 & \textbf{92.68} & \textbf{86.66} & 93.63 & \textbf{80.52} & \textbf{92.89} & \textbf{93.68} & \textbf{85.92} \\
    & 5 & 78.66 & 92.23 & 86.52 & \textbf{94.03} & \textbf{80.52} & 92.74 & 93.51 & 85.90 \\
    & 10 & 78.36 & 92.56 & 86.62 & 93.89 & 80.22 & 92.74 & 93.30 & 85.74 \\ \midrule
   \multicolumn{2}{c}{Mean$\pm$Std. (iSICE)}  & {78.4$\pm$0.2} & {92.5$\pm$0.2} & {86.6$\pm$0.1} & {93.9$\pm$0.2} & {80.4$\pm$0.2} & {92.8$\pm$0.1} & {93.5 $\pm$0.2} & {85.6 $\pm$ 0.1} \\
    \bottomrule
  \end{tabular}}
  \caption{Performance of iSICE on changing number of iterations ${N}$ while fixing sparsity constant $\lambda$ and learning rate $\eta$ to 0.01 and 1.0, respectively. The mean and standard deviation (std.) of the classification performance resulted by iSICE are also shown (only single precision is shown for ease of presentation) for a better understanding of number of iterations changes in iSICE. Results are shown on multiple datasets with VGG-16 and ResNet-50 backbones. The best results in each column (including those that surpass the performance of iSQRT-COV) are highlighted with boldface.}
  \label{tab: iterations change}
\end{table*}

\section{Training iSICE with Different Backbones}
\label{app:backb}
This section provides the settings of training different backbones with iSICE mentioned in Section \ref{exp:data} of the main text. We train iSICE with following backbones: VGG-16 \citelatex{simonyan2014very_supp}, ResNet-50 \citelatex{he2016deep_supp}, ConvNext-T \citelatex{liu2022convnet_supp}, and Swin-T \citelatex{liu2021swin_supp}. All backbones are pre-trained on ImageNet-1k \citelatex{deng2009imagenet_supp}. We use the pre-trained weights provided by the torchvision 0.13.0 package that comes with PyTorch library \citelatex{paszke2019pytorch_supp}.

All backbones produce more than 256 feature channels. In the recent literature on covariance representation \citelatex{li2017second_supp,li2018towards_supp,yu2018statistically_supp}, a common practice is to experiment with 256 channels for efficiency and compactness of final representation. To compare our method with the recent literature, we also conducted most of our experiments with 256 channels. Additionally, we conducted experiments with 512 channels to demonstrate the effectiveness of the proposed iSICE in working with a larger number of feature channels (provided in Table \ref{tab: sice size} in the main text). For reducing the original number of feature channels from 2048/512 to 256, we add a $1\times 1$ convolution layer, batch normalisation and ReLU activation layers after the last convolution layer (in case of CNN models) and transformer block (in case of Swin Transformer). We then compute iSICE with the reduced feature channels, and  we only use the upper-triangular entries of the symmetric matrix as a representation.

We fine-tune all backbones for 50-100 epochs with AdamW optimiser \citelatex{loshchilov2017decoupled_supp}. We fine-tune VGG-16 and ResNet-50 CNN backbones with an initial learning rate of 0.00012, and ConvNext-T CNN and Swin-T transformer backbones with an initial learning rate 0.00005. For all backbones, we decrease the learning rate by a factor of 10 at the 15th and 30th epochs. Depending on the dataset and backbones, our fine-tuning process lasts for about 3-8 hours with four P100 GPUs, 12 CPUs and 12GB memory. We provide the source code of our method as supplement material for reproducing the experiments.

\section{Robustness of iSICE to  Hyper-parameters}
\label{app:hyper}

This section includes the experiments mentioned in ``Robustness of iSICE on Hyper-parameter Changes'' of  Section \ref{sec:eval} of the main text. We have conducted comprehensive experiments with the hyper-parameter range mentioned in the main text to demonstrate the robustness of iSICE with respect to hyper-parameter changes. In Tables \ref{tab: comparison-with-all}, \ref{tab:in100} and \ref{tab: sice size} of the main text, we showed the results obtained by using a consistent hyper-parameter set across different backbones and datasets to avoid overfitting. Specifically, we showed the results obtained with the median (marked with boldface) of the hyper-parameter range, \ie, $\lambda$ = \{1.0, 0.5, 0.1, \textbf{0.01}, 0.001, 0.0001, 0.00001\}, $\eta$ = \{0.001, 0.01, 0.1, \textbf{1.0}, 5.0, 10.0, 20.0\}, and ${N}$ = \{1, \textbf{5}, 10\}.

Below we show some experiments to demonstrate the robustness of iSICE with respect to hyper-parameter changes. Specifically, we analyse the performance of iSICE when one hyper-parameter changes while the other two are fixed. For consistency, we experiment with the same hyper-parameters used for reporting the performance of iSICE across the tables of the main text, \ie, $\lambda$ = 0.01, $\eta$ = 1.0, and ${N}$ = 5. As mentioned above, we will fix two of them and vary the third one to observe its impact to the performance of the proposed iSICE. All of our experiments are compared with COV and Precision $\mathbf{\Omega}$ methods (please refer to the main text for details).

\vspace{0.1cm}
\noindent\textbf{Robustness against sparsity constant changes.} In this experiment, we change the sparsity constant $\lambda$ while keeping the learning rate $\eta$ and the number of iterations ${N}$ fixed. Our experimental results are shown in Table \ref{tab: sparsity change}. From the results, we can clearly see that across all datasets, the change of sparsity constants does not significantly impact the performance of iSICE. The VGG-16 based iSICE shows more robustness toward sparsity constant changes. The classification performance for the MIT dataset appears to have been more significantly impacted due to sparsity constant changes than the other three fine-grained datasets. Specifically, on the three fine-grained datasets, the standard deviation of results is significantly low, \ie, less than 0.51 when compared with the mean values ranging between 85.50 to 93.85. This confirms that our method can be used for fine-grained image classification purposes with a reasonable range of sparsity constant.

\vspace{0.1cm}
\noindent\textbf{Changing the learning rate.} In this experiment, we change the learning rate while keeping the sparsity constant and the number of iterations fixed. Our experimental results are shown in Table \ref{tab: lr change}. From the results, we can see that across all datasets, the change in learning rate does not significantly affect the performance. Two datasets, namely Birds and Cars, have shown less impact on performance, as suggested by the standard deviation of 0.14 or lower. The other two datasets also show a low standard deviation of results. The low standard deviation across a wide range of learning rates (from 0.001 to 20.0) shows that our method is robust to the changes in learning rate and a small learning rate such as 0.01 can be used for computing SICE with our method.

\vspace{0.1cm}
\noindent\textbf{Changing the number of iterations.} Below we change the number of iterations while keeping the sparsity constant and the learning rate fixed. Our experimental results are shown in Table \ref{tab: iterations change}. The results show that regardless of the CNN backbones used, the change in the number of iterations can vary the performance only up to 0.20. It is also noticeable that our method is able to give good performance even with two iterations only. This experiment shows that our method is not sensitive to the changes in the number of iterations.

\section{iSICE with Learning Rate and Sparsity Modulators}
This section provides additional experiments on iSICE with MLP modulators introduced in ``iSICE with learning rate and sparsity modulators'' of Section \ref{sec:eval} of the main text. In Table \ref{tab:beta-exp}, the CNN feature maps with average pooling were used to learn on-the-fly the learning rate and sparsity. In Table \ref{tab:beta-exp-2}, we provide an additional experiment with MLP when both $\nabla_1$ from Alg. \ref{alg:two} and average-pooled feature maps are used, \ie, concatenated before passing them into the modulator. The combination of $\nabla_1$ from Alg. \ref{alg:two} and average-pooled feature maps further improve the classification performance of iSICE.

\begin{table}[!htbp]
    \centering
    \setlength{\tabcolsep}{0.5em}
    \resizebox{1.0\linewidth}{!}{
    \begin{tabular}{lccccc}
    \toprule
     Method &  MIT & Airplane & Birds & Cars & ImageNet100\\ \midrule
     iSICE & 80.5 & 92.7 & 85.9 & 93.5 & 74.8\\
     \rowcolor{LightCyan}  iSICE+MLP & 81.3 & 93.4 & 86.1 & \textbf{93.9} & 76.3\\
     \rowcolor{bananamania}  iSICE+MLP$^*$ & \textbf{81.8} & \textbf{93.8} & \textbf{86.4} & \textbf{93.9} & \textbf{77.1}\\
     \bottomrule
    \end{tabular}
    }
    \caption{Comparison between the classification performance of iSICE, iSICE+MLP and iSICE+MLP$^*$ (improved variant) on ResNet-50. Results of iSICE+MLP in second row use $\textbf{X}$ to compute learning rate and sparsity with modulators. Results of iSICE+MLP$^*$ in the third row use a concatenation of both $\textbf{X}$ and $\Delta_1$ from Alg. \ref{alg:two} to compute on-the-fly learning rate and sparsity by MLP-based modulators.}
    \label{tab:beta-exp-2}
\end{table}

\section{Memory Consumption}
\label{app:memory comsumption}
This section describes memory consumption on iSICE mentioned. Algorithm \ref{alg:two} stores the $d\times{d}$ matrices of ${\mathbf \Sigma}$, ${\mathbf S}_i$, ${\mathbf S}^{+}_i$ and ${\mathbf S}^{-}_i$, \etc. The memory complexity of Algorithm \ref{alg:two}  is approximately $\mathcal{O}\big(d^2(N\!+\!N_s)\big)$, where $d$ denotes the channel size, $N$ is iSICE iterations, and $N_s$ is Newton-Schulz iterations. For typical $d=256$, $N=5$, $N_s=5$, iSICE uses approximately $3\times 10\times 8 \times 256^2 = 0.012$ GB memory which is a tiny fraction of memory that the backbone consumes.

\section{Visualisation of Learned Feature Maps}
\begin{figure}[h]
    \centering
    \includegraphics[width=1\linewidth]{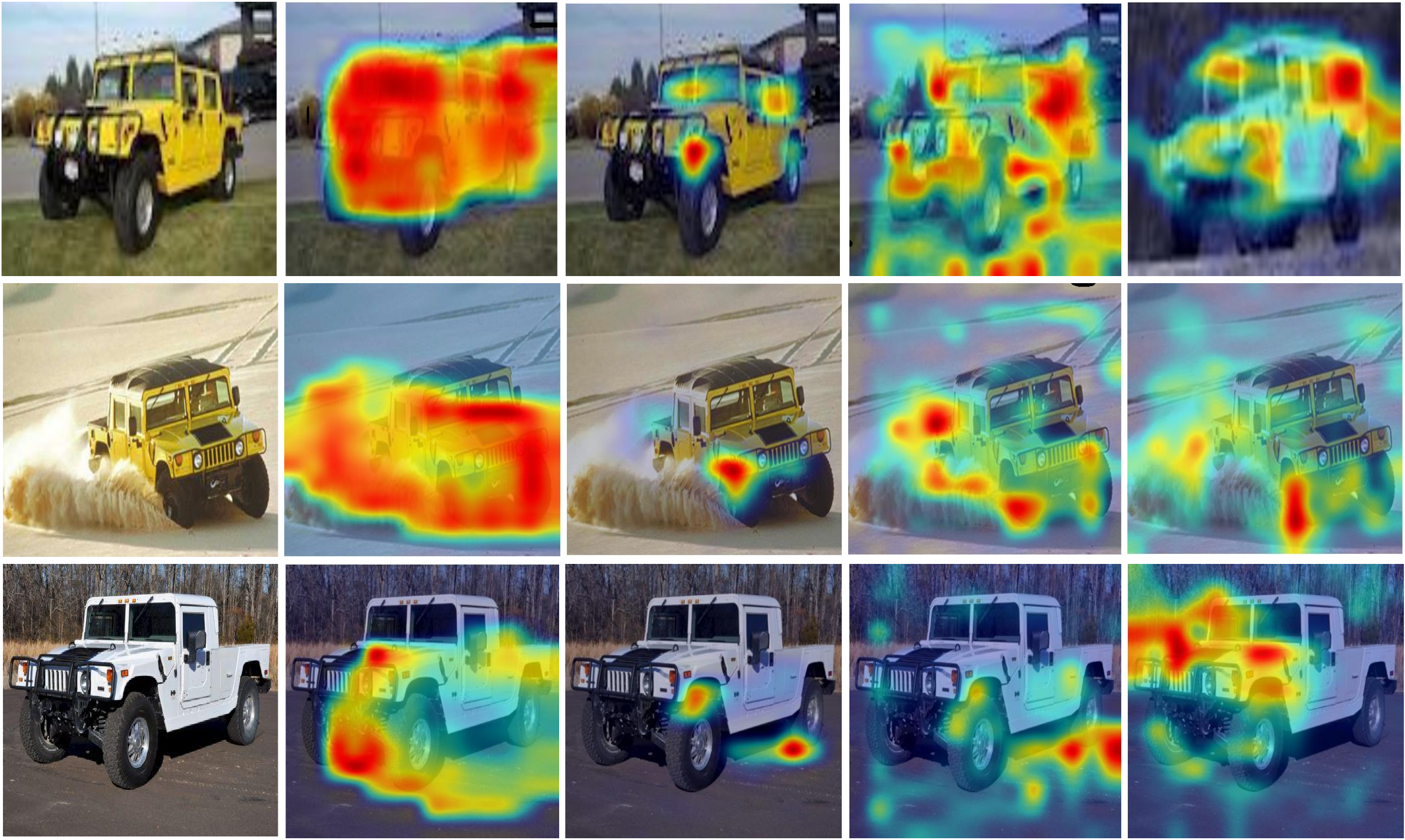}
    \vspace{-0.3cm}
    \caption{Visualisation of learned convolutional feature maps from Cars dataset with ResNeXt-101 backbone. From left: Input image, GAP, iSQRT-COV, Precision Matrix and iSICE.}
    \label{fig: heatmaps}
\end{figure}

Below we visualize the convolutional feature maps learned by the CNN model with different methods. We extract feature maps from the last convolution layer and perform average pooling on them. We convert the pooled feature map to a heatmap and draw it over the input image. The colour in the heatmap ranges from blue to red, blue indicates cold and red indicates hot. Fig.~\ref{fig: heatmaps} suggests that with iSICE, the model focuses well on the key parts of car to extract features for classification. GAP (global average pooling) overly focuses on entire foreground, iSQRT  focuses poorly, while iSICE lets us control the degree of `focus' by controlling sparsity.

{\small
\bibliographystylelatex{ieee_fullname}
\bibliographylatex{latex}
}

\end{document}